\documentclass[conference]{IEEEtran}
\IEEEoverridecommandlockouts

\usepackage [utf8] {inputenc}
\usepackage [T1]   {fontenc}

\usepackage{booktabs}
\usepackage[
  table,
  dvipsnames
]{xcolor}
\usepackage[pdftex]{graphicx}
\usepackage{subcaption}
\usepackage{array}
\usepackage{adjustbox}
\usepackage{rotating}

\graphicspath{{figures/}} 
\DeclareGraphicsExtensions{.pdf,.jpg,.png} 
\usepackage{mathtools,amsmath, amssymb} 

\usepackage{amsfonts} 
\usepackage{tikz} 
\usetikzlibrary{shapes,arrows,fit,calc,positioning,automata,backgrounds}
\usepackage{multirow}
\usepackage{hhline}
\usepackage[hidelinks]{hyperref}
\usepackage{enumerate}
\usepackage{epstopdf}
\usepackage{dsfont}
\usepackage{siunitx}
\usepackage{cite}
\usepackage{diagbox}

\colorlet{Green1}{green!90!}
\colorlet{Green2}{green!60!}
\colorlet{Green3}{green!40!}
\colorlet{Green4}{green!20!}
\colorlet{Green5}{green!10!}

\definecolor{Bookcolor}{HTML}{00F9DE}

\makeatletter
\def\@citex[#1]#2{\leavevmode
\let\@citea\@empty
\@cite{\@for\@citeb:=#2\do
{\@citea\def\@citea{,\penalty\@m\ }%
\edef\@citeb{\expandafter\@firstofone\@citeb\@empty}%
\if@filesw\immediate\write\@auxout{\string\citation{\@citeb}}\fi
\@ifundefined{b@\@citeb}{\hbox{\reset@font\bfseries ?}%
\G@refundefinedtrue
\@latex@warning
{Citation `\@citeb' on page \thepage \space undefined}}%
{\@cite@ofmt{\csname b@\@citeb\endcsname}}}}{#1}}
\makeatother

\usepackage{scalerel} 
\usepackage{tikz} 
\usetikzlibrary{svg.path} 
\definecolor{orcidlogocol}{HTML}{A6CE39}
\tikzset{
  orcidlogo/.pic={
    \fill[orcidlogocol] svg{M256,128c0,70.7-57.3,128-128,128C57.3,256,0,198.7,0,128C0,57.3,57.3,0,128,0C198.7,0,256,57.3,256,128z};
    \fill[white] svg{M86.3,186.2H70.9V79.1h15.4v48.4V186.2z}
                 svg{M108.9,79.1h41.6c39.6,0,57,28.3,57,53.6c0,27.5-21.5,53.6-56.8,53.6h-41.8V79.1z M124.3,172.4h24.5c34.9,0,42.9-26.5,42.9-39.7c0-21.5-13.7-39.7-43.7-39.7h-23.7V172.4z}
                 svg{M88.7,56.8c0,5.5-4.5,10.1-10.1,10.1c-5.6,0-10.1-4.6-10.1-10.1c0-5.6,4.5-10.1,10.1-10.1C84.2,46.7,88.7,51.3,88.7,56.8z};
  }
}

\newcommand\orcidicon[1]{\href{https://orcid.org/#1}{\mbox{\scalerel*{
\begin{tikzpicture}[yscale=-1,transform shape]
\pic{orcidlogo};
\end{tikzpicture}
}{|}}}}

\newcommand{\psnrSoneMzero}{29.3}
\newcommand{\psnrSoneMthree}{30.2}
\newcommand{\cdSfourMzero}{58}        
\newcommand{\cdSfourMthree}{369}     
\newcommand{\fmSoneMzero}{0.21}       
\newcommand{\fmSoneMthree}{0.37}
\newcommand{\ablDepthDrop}{31\%}      
\newcommand{\regRateClear}{99.5\,\%}  
\newcommand{\regRateMid}{1.0\,\%}     
\newcommand{\regRateHigh}{0.0\,\%}    
\newcommand{\poseGap}{0.13}           
\newcommand{\surfMzeroClear}{99}      
\newcommand{\trivMarginClear}{7.9}   
\newcommand{\trivMarginHigh}{3.4}    

\begin{document}
\title{Gaussian Splatting Underwater: A Controlled Cross-Regime Study\\}

\author{
Olaya~Álvarez-Tuñón \orcidicon{0000-0003-3581-9481},
Stella Graßhof \orcidicon{0000-0002-6791-7425}
}
\maketitle

\begin{figure}[!b]
\vspace{-0.2cm}
\noindent\rule{\columnwidth}{0.4pt}\\
\footnotesize Main funding source(s): the Innovation Fund Denmark, project DeepODO (Deep visual odometry for underwater intervention drones).

\footnotesize \copyright~2026 IEEE. Personal use of this material is permitted. Permission from IEEE must be obtained for all other uses, in any current or future media, including reprinting/republishing this material for advertising or promotional purposes, creating new collective works, for resale or redistribution to servers or lists, or reuse of any copyrighted component of this work in other works.
\end{figure}

\begin{abstract}
The underwater environment is challenging for 3D reconstruction, because particles suspended in the water scatter and diffuse light, turbidity varies,
absorption depends on wavelength, and illumination is rarely uniform. Methods based on Gaussian splatting have generally been developed for conditions that allow good image quality, and have primarily been tested on relatively shallow
water. This paper examines how well Gaussian splatting performs across publicly available underwater datasets representing different degrees
of turbidity, loss of illumination, and colour attenuation, together with an industrial survey. Five systems with public code are run under one protocol, with shared poses, initialisation, budget, and evaluator, to establish their
relative advantages, disadvantages, and limitations. What these methods can do turns out to depend more on the setup than on the architecture. Water clarity binds upstream of rendering, since structure-from-motion registers
\regRateClear{} of frames in clear water and \regRateHigh{} at 12\,NTU. Illumination geometry decides whether a medium model helps at all: under an artificial light that moves with the camera, medium-blind splatting beats both medium-aware systems. On the survey the benchmark's photometric leader comes last, beaten on geometry by a
restoration pre-pass in front of vanilla 3DGS --- and none of it is visible in the scores the field reports. Scene builds, per-run configurations, and evaluation code are released at \url{https://github.com/olayasturias/uw3dgs}.%
\end{abstract}



\section{Introduction}

Underwater 3D scene reconstruction is fundamental to marine robotics, with applications in subsea structural inspection, self-localisation for autonomous underwater vehicles, archaeological survey and environmental monitoring.

3D Gaussian splatting (3DGS)~\cite{kerbl20233dgaussiansplatt} achieves
photo-realistic, real-time rendering from explicit anisotropic primitives, and
recent work~\cite{li2024watersplatting,yang2024seasplat,wang2024uwgs} has
adapted it to underwater imagery. That work is developed and evaluated in
relatively shallow water or highly controlled datasets, and it is not known
whether it holds for real ROV operations in deeper, more turbid conditions,
where a render can look right while the geometry underneath it is wrong.

The present work asks: \emph{which of the available methods are solving
underwater 3D reconstruction, and which are only producing visually plausible
renders?} Underwater methods repair two failures at once, corrupted appearance
and geometry fabricated to explain the medium, and photometric metrics cannot
separate them. It is therefore not known which of the two a medium model buys,
or whether a plain restore-then-splat baseline already covers the first. We
answer with a controlled protocol rather than a new method: shared poses,
initialisation, budget and evaluator across five code-available systems and
four water regimes, with turbidity as a measured independent variable.

The paper contributes:
(1)~a cross-regime \emph{analysis} of five state-of-the-art methods under a common protocol, over four underwater regimes ranging from the field's default shallow benchmark, to a controlled tank, a deep site under co-moving light, and an industrial survey; \\
(2)~a component \emph{analysis} inside one of the method's codebases~\cite{yang2024seasplat} separating the medium model, the depth prior and the backscatter term;\\
(3)~a \emph{benchmark protocol} adding geometric evaluation in addition to the existing photometric axis,
including a reference-free floater-mass statistic that quantifies veil contamination where no ground truth exists; and \\
(4)~a measurement of how far a fitted medium model carries beyond the scene it
was fitted on, which the field assumes but does not test.

\begin{figure}[t]
    \centering
    \includegraphics[width=1\linewidth]{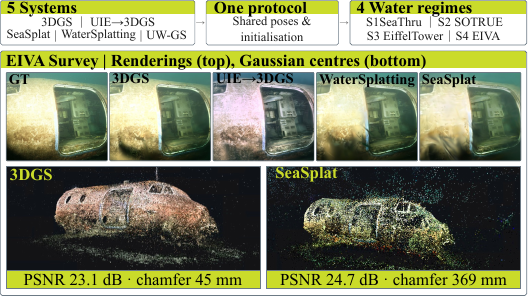}
    \caption{The proposed controlled cross-regime study, exemplified with EIVA Survey's renderings and point clouds. The point clouds correspond to the Gaussians' centres.}
    \label{fig:GA}
\end{figure}

Figure~\ref{fig:GA} exemplifies the central result: photometric and geometric quality move in opposite directions, and the metric
the field standardised on saturates precisely where the problem becomes hard.

\section{Background}
\subsection{Underwater image formation}
The underwater image formation model is commonly formulated in the literature as:
\begin{equation}
I_c \;=\; J_c\, e^{-\beta^D_c z} \;+\; B^\infty_c\!\left(1 - e^{-\beta^B_c z}\right),
\label{eq:uifm}
\end{equation}
with the first term modelling the directly transmitted signal, attenuated with
range, and the second the backscattered veil that accumulates along the same
path. $J_c$ is the true scene colour in channel $c$, $z$ the range, $\beta^D_c$
the wideband attenuation coefficient, $\beta^B_c$ the backscatter coefficient,
and $B^\infty_c$ the radiance of the background light at infinite
range~\cite{alvarez2019generation}.
Two properties of Eq.~\eqref{eq:uifm} drive every design decision in the
systems tested. First, $\beta^D \neq \beta^B$ and both depend on
wavelength~\cite{akkaynak2018arevisedunderwat,akkaynak2019seathru}, so
single-coefficient haze models imported from the dehazing literature are
physically wrong in water, an error invisible in grey fog and severe in
colour. Second, every term depends on range $z$: recovering colour and
recovering geometry are the same problem.

\subsection{3D Gaussian Splatting: State of the Art}
\label{sec:designspace}

3DGS owes its efficiency to its explicit representation: radiance lives on the
primitives, with no per-ray evaluation. Underwater, however, scattering is a
function of range $z$, and the only way such a representation can produce it
is to place semi-transparent primitives in the water
column~\cite{li2024watersplatting, liu2024aquaticgs}. The design space of
underwater 3DGS lies in three aspects: medium realisation, additional
supervision, and per-Gaussian state.

\textit{Medium realisation.} Vanilla 3DGS ignores the medium its primitives sit
in. Underwater variants place it in image space, as a restoration pre-pass
before splatting~\cite{shi2024aquags,qiao2025restorgs}; in a separate
volumetric field rendered jointly with the
splats~\cite{li2024watersplatting,liu2024aquaticgs,wu2025plenodium};
per-Gaussian, as analytic attenuation and backscatter terms on each
primitive~\cite{yang2024seasplat,wang2024uwgs,yuan20253duir,li2025dualphysgs,zhang2026watercleargs};
or inside the compositing rule
itself~\cite{mualem2024gaussiansplashin}. The choice fixes the failure mode: a
pre-pass restores each view independently and cannot enforce multi-view
consistency, whereas a separate field reintroduces the per-ray evaluation that
3DGS exists to avoid.

\textit{Additional supervision.} Because the medium couples colour to range,
most methods add constraints on geometry: monocular depth
supervision~\cite{yang2024seasplat,du2024udrgs}, density control that keeps
attenuation from starving far-field
densification~\cite{wang2024uwgs}, masks for moving distractors such as fish
and suspended particles~\cite{wang2024uwgs,gough2025aquanerf}, and semantic
guidance~\cite{jiang2025semanticguidedga}.

\textit{Per-Gaussian state.} A vanilla primitive carries a position, a
covariance factorised into scale and rotation, an opacity, and a view-dependent
colour as degree-3 spherical harmonics (SH). SeaSplat drops to SH degree~0, so
appearance variation must come from its medium networks or from
geometry~\cite{yang2024seasplat}, while UW-GS~\cite{wang2024uwgs} and
SeaFree-GS~\cite{liu2025seafreegs} carry a clean and a degraded colour per
primitive and map between them with a network.

\begin{table*}[t]
\caption{System anatomy. All systems initialise from the shared per-scene SfM cloud
of Sec.~\ref{sec:protocol}. SH = spherical harmonics. Entries in \textbf{bold} are
undocumented in the original paper.}
\label{tab:anatomy}
\centering\scriptsize\setlength{\tabcolsep}{3pt}
\begin{tabular}{@{}p{2.5cm} p{3.0cm} p{5.0cm} p{5.4cm}@{}}
\toprule
 & Per-Gaussian state & Medium realisation & Extra supervision / inputs \\
\midrule
M0 vanilla 3DGS~\cite{kerbl20233dgaussiansplatt} & pos., SH-3 colour, opacity, scale, rot. & --- &
--- \\

M1 UIE$\rightarrow$3DGS~\cite{kerbl20233dgaussiansplatt} & as M0 & removed in 2D pre-pass: gray world + CLAHE &
restored images as input \\

M2 WaterSplatting~\cite{li2024watersplatting} & as M0 (gsplat param.) &
separate field: dir.-conditioned MLP $\rightarrow$ medium colour, attenuation and backscatter densities &
medium regularisers \\

M3 SeaSplat~\cite{yang2024seasplat} & pos., \textbf{SH-0 colour}, opacity, scale, rot. &
analytic Eq.~\eqref{eq:uifm} in image space; CNNs infer $\beta^D\!, B^\infty$ from rendered depth &
depth-smooth, depth-weighted recon., dark-channel, grey-world \\

M4 UW-GS~\cite{wang2024uwgs} & as M0 + clean/degraded colour pair via view--distance MLP &
per-Gaussian distance-dependent transform &
depth-alignment loss; \textbf{mono-depth maps required} \\
\bottomrule
\end{tabular}
\end{table*}
\section{Experimental Design}

\subsection{Scenes}
Four regimes, labelled S1--S4, run from experimental control to realistic
operating conditions.
\textbf{SeaThru-NeRF Cura\c{c}ao (S1)}~\cite{levy2023seathrunerf} is the
in-distribution control, since most surveyed methods were tuned on this
benchmark family. We use the authors' own COLMAP reconstruction.
\textbf{SOTRUE (S2)}~\cite{apl2025sotrue} provides six turbidity levels
measured with a Seapoint meter in Nephelometric Turbidity Units (NTU), each
captured along an identical servo-driven trajectory with poses from motor
encoders rather than from the images, so turbidity is the only variable that
changes between sequences. Its monochrome imagery switches off the
wavelength-dependent half of Eq.~\eqref{eq:uifm} and leaves the
range-dependent half in isolation.
\textbf{Eiffel Tower (S3)}~\cite{boittiaux2022eiffeltower} was captured in deep
water, where the uniform-veiling-light assumption is structurally violated by
the artificial light source moving with the ROV.
\textbf{EIVA survey (S4)} is a proprietary operational survey with a
photogrammetric reference point cloud.

\subsection{Methods}
Five methods are evaluated here, one for each cell of the medium-realisation axis of
Sec.~\ref{sec:designspace}: no medium model (\textbf{M0}, vanilla
3DGS~\cite{kerbl20233dgaussiansplatt}); a medium removed in image space
(\textbf{M1}, a classical enhancement pre-pass in front of that same vanilla
3DGS); a medium held in a separate field (\textbf{M2},
WaterSplatting~\cite{li2024watersplatting}); and a medium attached to the
primitives (\textbf{M3}, SeaSplat~\cite{yang2024seasplat}, and \textbf{M4},
UW-GS~\cite{wang2024uwgs}). Three further systems we built and smoke-tested
duplicate cells already covered and are not reported.
Table~\ref{tab:anatomy} states what each system changes, read from its
released code rather than from its paper.

M1's pre-pass is gray-world white balance~\cite{buchsbaum1980grayworld}
followed by CLAHE~\cite{zuiderveld1994clahe} on the lightness channel, clip
2.0 and $8\times8$ tiles, applied identically to every scene. It is the
classical core of fusion-based underwater
enhancement~\cite{ancuti2018colorbalance} without the fusion stage: no
training data and no network, so it is the cheapest conceivable underwater
adaptation and the baseline the medium-aware methods have to beat.

\subsection{Controlled protocol}
\label{sec:protocol}
Published underwater-splatting numbers cannot be compared with each other, since
they differ in poses, initialisation, splits, resolution, and measurement
metrics at once. Our protocol removes those differences so that any remaining
gap between systems is attributable to the systems themselves.

\textbf{Poses.} The same per scene for every system, and no system runs its own
SfM. They come from motor encoders on S2 and from the datasets' own
reconstructions elsewhere. How often COLMAP succeeds unaided is measured
separately and reported as a result.

\textbf{Initialisation.} Every system starts from the same sparse cloud per
scene, never taken from the evaluation reference: the dataset's own on S1 and
S3, and our own triangulation with poses held fixed on S2 and S4 (1.77 and
0.34\,px mean reprojection error). On S2 the clear-water cloud seeds every
turbidity level, since from 7\,NTU upward no usable triangulation exists,
which is itself a result.

\textbf{Budget.} One run per cell, 30\,k optimisation iterations at each
system's default densification schedule, on a single 24\,GB GPU. Where a
default schedule exceeds that budget we report the outcome rather than tuning
it away.

\textbf{Repeats.} Two cells at 7\,NTU on S2 were repeated to bound run-to-run
spread. Running M0 three times at a fixed configuration gives surface errors
of 848, 844 and 846\,mm, so the nondeterminism of the CUDA reductions is worth
roughly $\pm2$\,mm. Running M2 at three seeds gives 660, 578 and 581\,mm, a
spread of 82\,mm, which is the figure to read the S2 depth column against.
Elsewhere we use the uncertainty matched to the measurement: on S4 that is the
ICP residual of the alignment the chamfer is computed through. Differences
below either are treated as ties rather than rankings.

\textbf{Splits.} Every 8th frame by sorted filename, the same index set for
every system, leaving 3 held-out views on S1, 25 on S2 and 71 on S4.
Photometric scores use the in-medium render rather than any restored output,
so systems producing both are compared on the same quantity. M1 is the
exception, since its targets are its own enhanced images; its photometric
cells are marked and are not comparable to the rest.

\textbf{Resolution and inputs.} Native resolution with the forks' automatic
downscaling disabled, except on S4, whose $1408\times1408$ frames are a
$0.5\times$ resize of the rectified originals applied once before any system
sees them. Inputs are identical per scene, with 32 border pixels masked on S2
to exclude its documented edge distortion.


\subsection{Metrics}\label{sec:metrics}

\textbf{Photometric.} PSNR, SSIM~\cite{wang2004ssim} and LPIPS~\cite{zhang2018lpips} on held-out
views: per-pixel fidelity, local structural similarity, and distance in the
feature space of a pretrained network.

\textbf{Geometry-free control.} A photometric score alone cannot separate a
good reconstruction from an easy image, so we also score a baseline with no
geometry at all: the per-pixel median of a scene's training frames, one static
image, evaluated against the same held-out views with the same code. Whatever a
system scores above it is what reconstruction earns, and its own behaviour
across turbidity separates the metric from the method.

\textbf{Geometric.} On S2 we compare rendered depth against an independent
stereo reference. That depth is an alpha blend and translucent matter dominates
it, so a model wrapped in haze reports the haze's depth rather than the
surface's. We therefore render depth twice, once as the plain blend and once
over only the primitives with opacity above 0.5, and the gap between them
measures how much floating matter the model committed to. On S4 we compare
opacity-gated Gaussian centres against the photogrammetric reference cloud
(accuracy, completeness, chamfer) on the co-visible region, after an ICP
alignment~\cite{besl1992icp} whose residual we report. Where no geometric
reference exists we instead report
\emph{floater mass}, the fraction of total opacity sitting more than $\tau$
nearer the camera than the rendered surface, at $\tau=0.1$\,m unless a table
says otherwise (Table~\ref{tab:abl} sweeps it). It counts translucent mass in
front of the surface and is blind to opaque mass in the wrong place, a
distinction S4 makes visible (Sec.~\ref{sec:e1e5}).

\textbf{Registration.} The fraction of frames COLMAP~\cite{schoenberger2016sfm}
registers. We report it on
S2 alone, the one scene whose sequences differ in turbidity only and whose
poses come from outside the images, so the rate isolates turbidity.

\subsection{The experiments}
Six experiments run over those scenes. \textbf{E1} compares all five systems
on the benchmark (S1) and the operational survey (S4). \textbf{E2} sweeps four
measured turbidity levels on S2 with trajectory, poses and initialisation
fixed. \textbf{E3} retrains M3 on S2 at 0\,NTU with COLMAP poses in place of
encoder ground truth, and only M3, since it is the only system whose losses
read geometry back out of its own render (Table~\ref{tab:anatomy}) and so the
one most exposed to pose error; above 0\,NTU COLMAP registers almost no frames,
which is itself the result. \textbf{E4} switches SeaSplat's medium model, depth
prior and backscatter term off one at a time on S1. \textbf{E5} covers S3, and
additionally runs M0 on an every-16th-frame version of the same dive to
separate view overlap from regime. \textbf{E6} renders the S3 model through the
medium networks fitted on S1, an evaluation with no training run.

Two systems carry requirements that bound their coverage. M1 restores colour
before splatting, which is close to meaningless on the monochrome S2, so it was
not attempted there or on S3. M4 needs a monocular depth map per input image,
which we prepared for S1 and S4 only.

\section{Results}

\begin{table}[tb]
\caption{E2 turbidity sweep and E3 pose source on SOTRUE. \emph{Ctrl} is the geometry-free control of Sec.~\ref{sec:metrics}. The indented \emph{$+$ COLMAP poses} rows replace the
encoder poses with a free COLMAP reconstruction (E3).
$^{\dagger}$SeaSplat's default densification exceeds 24\,GB at 7\,NTU; that cell
alone uses a doubled threshold.}
\label{tab:e2turb}
\centering\scriptsize\setlength{\tabcolsep}{3pt}
\begin{tabular}{@{}lclcccc@{}}
\toprule
NTU & Ctrl$\uparrow$ & System & PSNR$\uparrow$ & Err$\downarrow$[mm] & Floater$\downarrow$ & \#G (k) \\
\midrule
\multirow{4}{*}{0.0} & \multirow{4}{*}{24.1}
 & M0 3DGS          & 31.97 & \textbf{99}  & 0.071 & 117 \\
& & M1 UIE$\rightarrow$3DGS & 26.85$^{*}$ & \textbf{68} & 0.058 & 296 \\
 & & M2 WaterSplatting& 32.87 & 289 & \textbf{0.001} & 154 \\
 & & M3 SeaSplat      & 28.17 & 438 & 0.047 & 269 \\
 & & \quad + COLMAP poses & 28.30 & --- & --- & 129 \\

\midrule
\multirow{2}{*}{6.0} & \multirow{2}{*}{31.5}
 & M0 3DGS          & 35.54 & 845 & 0.076 & 22 \\
& & M1 UIE$\rightarrow$3DGS & 29.83$^{*}$ & 531 & 0.177 & 24 \\
 & & M2 WaterSplatting& \textbf{35.59} & \textbf{456} & \textbf{0.000} & 18 \\
\midrule
\multirow{4}{*}{7.0} & \multirow{4}{*}{32.0}
 & M0 3DGS          & \textbf{35.94} & 848 & 0.068 & 23 \\
& & M1 UIE$\rightarrow$3DGS & 29.39$^{*}$ & 809 & 0.126 & 22 \\
 & & M2 WaterSplatting& 35.90 & \textbf{660} & \textbf{0.000} & 13 \\
 & & M3 SeaSplat      & 28.67$^{\dagger}$ & 934$^{\dagger}$ & 0.021$^{\dagger}$ & 111$^{\dagger}$ \\
 & & \quad + COLMAP poses & \multicolumn{4}{l}{\emph{\regRateMid{} registered --- no usable poses}} \\
\midrule
\multirow{4}{*}{12.0} & \multirow{4}{*}{31.7}
 & M0 3DGS          & 35.11 & 847 & 0.088 & 32 \\
& & M1 UIE$\rightarrow$3DGS & 29.58$^{*}$ & 861 & 0.082 & 30 \\
 & & M2 WaterSplatting& \textbf{36.38} & \textbf{809} & \textbf{0.002} & 3.1 \\
 & & M3 SeaSplat      & 28.13 & 917 & 0.049 & 282 \\
 & & \quad + COLMAP poses & \multicolumn{4}{l}{\emph{\regRateHigh{} registered --- no poses exist}} \\
\bottomrule
\end{tabular}
\end{table}

\begin{figure}[t]
\centering
\includegraphics[width=.7\columnwidth]{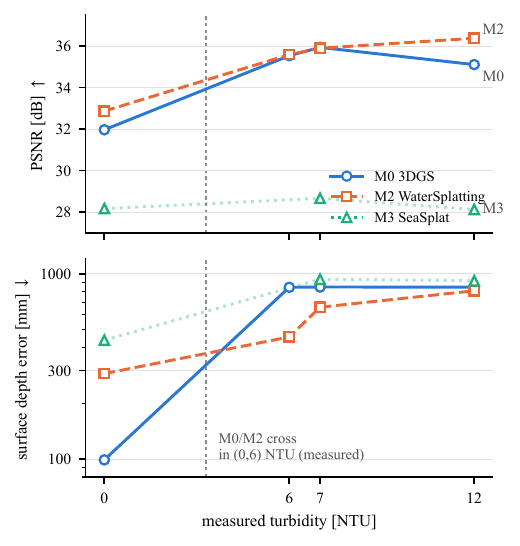}
\caption{Turbidity dose--response on SOTRUE (E2), for the three systems that
run there: M0, M2 and M3, the last without a 6\,NTU cell. PSNR (top) rises
with turbidity while surface depth
error against the stereo reference (bottom, log scale) grows by up to
$8.5\times$.}
\label{fig:dose}
\end{figure}

\begin{table*}[tb]
\caption{E1 cross-regime comparison, with the S3 deep-water probe (E5) below.
Chamfer is against each scene's photogrammetric reference; S1 has none, so
geometry there is floater mass. $^{*}$M1 is scored against its own enhanced
targets and is not comparable in-medium.}
\label{tab:e1}
\centering\footnotesize
\begin{tabular}{@{}llccccccc@{}}
\toprule
Scene & System & PSNR$\uparrow$ & SSIM$\uparrow$ & LPIPS$\downarrow$ &
Chamfer$\downarrow$ & Floater mass$\downarrow$ & Train (min) & \#G (M) \\
\midrule
\multirow{5}{*}{S1 Cura\c{c}ao}
 & M0 3DGS             & \psnrSoneMzero & 0.877 & 0.199 & n/a & \fmSoneMzero & 24 & 0.58 \\
 & M1 UIE$\rightarrow$3DGS & 25.0$^{*}$ & 0.807$^{*}$ & 0.251$^{*}$ & n/a & 0.191 & 24 & 1.34 \\
 & M2 WaterSplatting   & 29.5 & \textbf{0.919} & \textbf{0.141} & n/a & \textbf{0.077} & 40 & 1.06 \\
 & M3 SeaSplat         & \psnrSoneMthree & 0.895 & 0.173 & n/a & \fmSoneMthree & 82 & 3.35 \\
 & M4 UW-GS            & \textbf{30.6} & 0.898 & 0.179 & n/a & 0.551 & 188 & 0.93 \\
\midrule
\multirow{5}{*}{S4 EIVA}
 & M0 3DGS             & \textbf{25.9} & 0.780 & 0.509 & \cdSfourMzero  & 0.024 & 28 & 0.24 \\
 & M1 UIE$\rightarrow$3DGS & 23.1$^{*}$ & 0.577$^{*}$ & 0.665$^{*}$ & \textbf{45} & 0.015 & 27 & 0.38 \\
 & M2 WaterSplatting   & 25.7 & \textbf{0.783} & \textbf{0.494} & \textbf{46} & \textbf{0.001} & 34 & 0.09 \\
 & M3 SeaSplat         & 24.7 & 0.744 & 0.583 & \cdSfourMthree & 0.003 & 353 & 0.16 \\
 & M4 UW-GS            & \multicolumn{7}{l}{\emph{terminated at 19k/30k iterations after 13.5\,h ($\sim$30$\times$ slower than peers)}} \\
\midrule
\multirow{3}{*}{S3 Eiffel T.\ \emph{(probe)}}
 & M0 3DGS     & \textbf{27.0} & \textbf{0.891} & \textbf{0.193} & --- & 0.116 & 41 & 1.91 \\
  & M1 UIE$\rightarrow$3DGS & 24.4$^{*}$ & 0.833$^{*}$ & 0.291$^{*}$ & --- & 0.038 & 49 & 3.25 \\
 & M2 WaterSplatting & 23.2 & 0.731 & 0.483 & --- & 0.014 & 29 & \textbf{0.03} \\
 & M3 SeaSplat & 25.9 & 0.870 & 0.227 & --- & \textbf{0.008} & 132 & 1.99 \\
\bottomrule
\end{tabular}
\end{table*}

\begin{figure*}[ht]
\centering
\includegraphics[width=\textwidth]{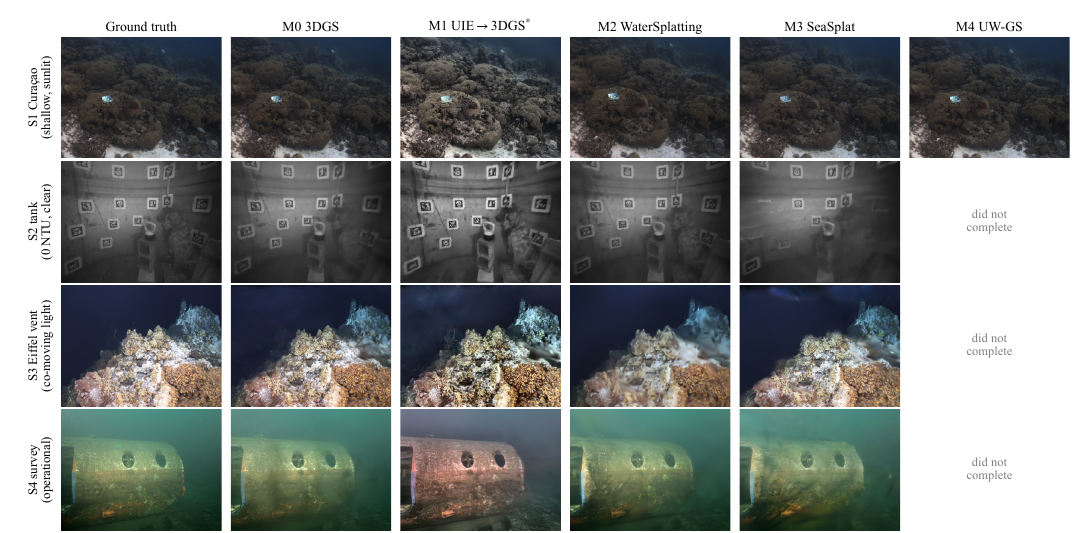}
  \caption{Renders of one held-out view per regime (rows) for ground truth and all five systems (columns). In the clear tank (S2) the AprilTags act as a sharpness scale and the visual ordering follows the measured surface error of
  Table~\ref{tab:e2turb}: M1 and M0 hold the tags, M2 softens them, M3 washes out the right half of the frame. Under co-moving light (S3) M2's render is the blurred residue of a 27\,k-Gaussian model after its medium field absorbed the scene; on the survey (S4) M3 is visibly hazier. $^{*}$M1 renders its own enhanced images, so its column
  differs from ground truth by construction; M4 completes only on S1.}
\label{fig:qual}
\end{figure*}

\subsection{E2: Turbidity Dose-Response}
\label{sec:e2}
Table~\ref{tab:e2turb} and Fig.~\ref{fig:dose} report the controlled sweep, in
which measured turbidity is the only variable. Two results inform what follows.

First, \textbf{the photometric and geometric axes move in opposite
directions}. For the medium-blind baseline (M0), held-out PSNR \emph{improves}
from 32.0\,dB in clear water to 35.9\,dB at 7\,NTU while its surface depth
error against the stereo reference degrades from \surfMzeroClear\,mm to
848\,mm: the veil leaves less texture to fit, and the optimiser reconstructs
the haze as geometry. The geometry-free control rises from 24.1 to 32.0\,dB
over the same range, so turbidity inflates PSNR for anything at all, and every
system's margin over that static image \emph{shrinks} as the water worsens
(M0: \trivMarginClear{} $\to$ \trivMarginHigh\,dB).

Second, \textbf{the systems fail geometrically in different shapes}. M0
collapses somewhere below 6\,NTU and then saturates near 850\,mm: once the
veil is reconstructed as a solid, no scene structure is left to lose. M2
degrades gradually, its scene absorbed into the medium field, shown by the
$50\times$ collapse in primitive count across the sweep. M3 fights the veil by densifying without bound, so
its runs either exhaust memory or saturate alongside M0. All three end at
$0.8$--$0.9$\,m while posting their best or near-best PSNR.

\subsection{E3: Pose Source}
\label{sec:e3}
With identical system, scene, and turbidity, swapping encoder ground-truth poses for COLMAP poses changes M3's PSNR by \poseGap\,dB (28.30 vs 28.17, the two M3
rows of the 0\,NTU block in Table~\ref{tab:e2turb}), within run-to-run noise. Where registration succeeds, pose \emph{accuracy} contributes essentially nothing to apparent quality. Registration
\emph{existence} is another matter. On SOTRUE's identical trajectories, COLMAP
registers \regRateClear{} of frames in clear water, \regRateMid{} at 7\,NTU (a
single two-view pair), and \regRateHigh{} at 12\,NTU, where not one match
survives geometric verification despite a detector tuned to fire on $10^4$
low-contrast features per image. These numbers characterise classical SIFT-based SfM, the pipeline the surveyed methods actually use; whether learned matchers survive the cliff is an open question
this study does not test.

\subsection{E1 and E5: Cross-Regime and Deep Water}
\label{sec:e1e5}
Figure~\ref{fig:qual} shows one example view per method;
Table~\ref{tab:e1} gives the numbers.

\textbf{On the field's benchmark, the image scores do not separate the systems, but the floater measure does.} All four in-medium systems land within 1.3\,dB of
each other on S1 (M0 at \psnrSoneMzero{}\,dB, M4 at 30.6\,dB), a tie by the
standards of this literature. However, a third of M3's Gaussians, and over half of M4's Gaussians, correspond to floaters, while M2 runs only with 0.077 floater mass.

\textbf{On the operational survey the ordering changes.} S4 has a
photogrammetric reference, so geometry is measured rather than inferred:
chamfer is the average separation between the reconstruction and that cloud.
M0 and M2 tie on appearance (25.9 vs 25.7\,dB) and both land close to the
reference (chamfer \cdSfourMzero{} and 46\,mm, ICP residual under 12\,mm). M3,
which led the benchmark, comes last on \emph{both} axes: 24.7\,dB and
\cdSfourMthree\,mm, its ICP fitness of 0.71 meaning 29\,\% of its committed
opacity has no counterpart on the reference surface. Its floater mass is low
(0.003), so that stray mass is solid geometry in the wrong place, not haze. It
is also the slowest by far, 353\,min against 28 on the same 566 views. That
case calibrates the floater statistic, S4 being the one scene with both axes:
floater mass ranks M3 second of four while chamfer ranks it last, so it
registers veil and misses displaced opaque geometry. LPIPS tracks SSIM throughout
Table~\ref{tab:e1} and is reported for completeness.

The naive baseline sits in that same group: M1's chamfer of 45\,mm lands
alongside M2's 46 and M0's \cdSfourMzero{} after 27\,min of training. Those
differences, 1 to 13\,mm, sit at or below the 12\,mm alignment residual they
are measured through, so we read the three as tied; chamfer is unaffected by
M1's enhanced targets, which bear only on its photometric cells.
This evidences that in operational surveys, a classic image enhancement beats the medium-aware architectures in precision and computational cost.

\textbf{Deep water with a moving light (E5, S3).} The two axes split as they
did in the tank. Vanilla 3DGS wins on appearance, by 1.1\,dB over SeaSplat
(27.0 vs 25.9\,dB, at about 1.9\,M Gaussians each) and 3.8\,dB over
WaterSplatting (23.2\,dB), and it carries the most veil of the three, 0.116
floater mass against 0.014 and 0.008. Whoever looks best floats most, as on
the benchmark. The medium models lose the appearance advantage they are built
for and keep the water out of the model. WaterSplatting loses it as in the
tank: its medium field is conditioned on viewing direction, a light
tied to the camera makes the illumination look directional, and the field
absorbs the light and most of the scene with it, leaving 27\,k Gaussians where
the others build 1.9\,M. 

The same M0 on a sparser version of the dive, every 16th frame, scores
16.5\,dB instead of 27.0: view overlap is worth 10.5\,dB, more than every
difference between methods above.

\begin{table}[t]
\caption{E4: component ablation inside SeaSplat (M3) on S1. A0$\rightarrow$A3 is the
total underwater contribution; A0$\rightarrow$A2 isolates the depth prior;
A0$\rightarrow$A1 isolates the backscatter term. S1 carries no metric geometry
reference, so the geometric axis is the GT-free floater mass of Sec.~\ref{sec:metrics}.}
\label{tab:abl}
\centering\scriptsize\setlength{\tabcolsep}{2.5pt}
\begin{tabular}{@{}lccccccc@{}}
\toprule
Config & Med. & Dep. & Bks. & PSNR$\uparrow$ & SSIM$\uparrow$ & Floater$\downarrow$ $\tau{=}.05/.1/.2$ & \#G\,(M) \\
\midrule
A0 full        & \checkmark & \checkmark & \checkmark & \textbf{30.22} & \textbf{0.895} & .41 / .37 / .33 & 3.35 \\
A1 $-$backsc.  & \checkmark & \checkmark & --         & 19.84 & 0.825 & .31 / .26 / .21 & 1.30 \\
A2 $-$depth    & \checkmark & --         & \checkmark & 27.46 & 0.881 & .30 / .25 / .21 & \textbf{0.62} \\
A3 $-$medium   & --         & \checkmark & --         & 23.98 & 0.865 & \textbf{.29 / .24 / .21} & 1.58 \\
\bottomrule
\end{tabular}
\end{table}

\subsection{E4: Component Ablation}
\label{sec:e4}

Table~\ref{tab:abl} switches SeaSplat's components off one at a time on S1.
Turning off backscatter (A1) costs 10.4\,dB of PSNR (30.2 to 19.8) but almost
no SSIM (0.895 to 0.825), because the error is one of brightness rather than
of structure: without a backscatter term the model cannot draw the haze, so
its render comes out too dark over most of the frame while the scene inside it
stays put. PSNR is a mean squared error and punishes that; SSIM compares local
contrast and barely notices.

The depth prior does the opposite of what it is meant to do. Turning depth
supervision off (A2) costs 2.8\,dB and the geometry gets \emph{better}:
floater mass falls from \fmSoneMthree{} to 0.25 (\ablDepthDrop{} less) on
0.62\,M Gaussians instead of 3.35\,M. These losses are supposed to pull mass
onto surfaces, but the cheapest way to satisfy them is to hang thin
semi-transparent sheets in front of the surface, so that is what the optimiser
builds. The appearance costs do not add up either: removing backscatter alone
costs $10.4$\,dB against $6.2$\,dB for removing the whole medium, since
attenuation without its veil term darkens the render further than no medium at
all. S1 has no metric geometry reference, so geometry here is floater mass
alone.

\subsection{E6: Cross-Site Medium Transfer}
\label{sec:e6}
The medium models fit one set of parameters per scene, which assumes the water
is optically uniform within it and, implicitly, that those parameters mean
something outside it. We tested the second half directly, freezing the medium
networks fitted on S1 (shallow, sunlit) and rendering the S3 model (deep,
artificial light) through them. PSNR falls from 25.9 to 17.0\,dB, a
$-9.0$\,dB penalty. The parameters this family estimates are therefore
site-specific: they describe the training images, not the water, so a
deployment cannot calibrate once and reuse.

\section{Discussion}

 \textbf{Parameters outside the primitives (M2) buy cleanliness and graceful
  degradation.} The separate field gives the cleanest geometry in the study
  (0.077 on S1, 0.001 on S4, $\leq$0.002 across the sweep) and degrades as a slope, not a cliff (289\,$\to$\,456\,$\to$\,660\,$\to$\,809\,mm). But because the field is a competing explanation for appearance, it can over-discriminate as field scene elements that change across frames with the moving light, as seen in S3.

  Per-Gaussian analytic medium (M3) buys appearance, paid for directly in geometry. As shown by the ablation study, removing the medium composite costs 6.2 dB, and removing just the backscatter term costs 10.4 dB, but the full model carries the worst floater mass of the entire ablation (0.368, versus 0.244–0.259 for every single-component removal). With nowhere else to put the veil, it manufactures translucent mass.

\textbf{Depth supervision buys 2.8\,dB and harms geometry.} Removing it
  improves floater mass (0.368\,$\to$\,0.253) and collapses the model
  (3.35\,M\,$\to$\,0.62\,M). The prior acts on the model's own rendered depth,
  and a smooth depth field is cheaper to obtain by spreading opacity across
  semi-transparent layers than by committing to one correct surface.

  A fixed image-space pre-pass (M1) buys the best geometry in the study by declining to model the medium at all. 68 mm on the clear tank (against M0's 99), 45 mm chamfer on the operational survey (best overall), 0.038 floater mass on the deep vent (cleanest real model there), and 27 minutes. Mechanically it works by treating the symptom: local contrast enhancement restores gradient signal on real structure, so
  points commit to surfaces. The price is that colour gets arbitrarily corrected by the enhancer, which makes the photometrics not comparable.

  Distractor handling plus modified densification (M4) buys benchmark rank and nothing else that survives contact with scale. Best S1 PSNR (30.6) but with the dirtiest field (0.551), 188 min on a 21-view scene, and infeasible beyond that.

  The decisive cost boundary is per-Gaussian evaluation. Systems whose medium lives outside the primitives train in 24–70 min across all four scenes; attaching the medium to the primitives costs 67 min–13.1 h, up to 12× on the same scene (S4: 28 min for M0 against 353 for M3), and M4 does not finish at all beyond the 21-view benchmark. Within the cheap band, cost tracks primitive count rather than coupling: M1 is the slowest of the three on S1 precisely because contrast enhancement supplies more gradient signal and drives densification to 1.34 M primitives.

\section{Conclusion}
Under one protocol across four water regimes, the outcome is set as much by deployment conditions as by architecture. The underwater-specific machinery buys appearance, but no geometry: on the operational survey, the best surfaces come from a fixed restoration pre-pass in front of vanilla 3DGS, which no medium-aware system improves upon.

Water clarity
settles the outcome before the renderer is reached, removing the poses above
roughly 7\,NTU; a moving light costs the medium models the advantage they
exist for, and what they estimate does not transfer between sites. 
Under harsh underwater imaging conditions, a reliable comparison of methods requires evaluating geometry alongside appearance, as the latter alone can be misleading.

The practical consequence is that appearance cannot carry a comparison in this domain.
On the field's own benchmark, photometric rank correlates \emph{positively} with contaminated geometry.
Comparing underwater reconstruction methods reliably requires a geometric axis reported alongside the photometric one. Currently, the field benchmarks itself almost exclusively on clear water, so the conditions where the problem is actually hard never appear in the evaluation.

\bibliographystyle{ieeetr}
\bibliography{references}

\end{document}